\documentclass{article}

  \usepackage[preprint]{neurips_2026}

\usepackage[utf8]{inputenc} 
\usepackage[T1]{fontenc}    
\usepackage{hyperref}       
\usepackage{url}            
\usepackage{booktabs}       
\usepackage{amsfonts}       
\usepackage{nicefrac}       
\usepackage{microtype}      
\usepackage{xcolor}         
\usepackage{multirow}
\usepackage{graphicx}
\usepackage{amsmath}
\title{Mamba-VGGT: Persistent Long-Sequence Video Geometry Grounded Transformer via External Sliding Window  Mamba Memory}

\author{%
Tianchen Deng\textsuperscript{1,4}, Zhenxiang Xiong\textsuperscript{1},
Nailin Wang\textsuperscript{1}
Fangjinhua Wang\textsuperscript{2}, Jiuming Liu\textsuperscript{3}, \\ \textbf{Jianfei Yang} \textsuperscript{4},
\textbf{Hesheng Wang\textsuperscript{1}},  \\
{\textsuperscript{\rm 1} Shanghai Jiao Tong University}
{\textsuperscript{\rm 2} ETH Zurich}
{\textsuperscript{\rm 3} Cambridge University} \\
{\textsuperscript{\rm 4} Nanyang Technological University}
}

\begin{document}

\maketitle


\footnotetext[1]{The first two authors contributed equally to this work.}

\begin{abstract}
  Visual Geometry Grounded Transformers (VGGT) have set new benchmarks in high-fidelity 3D scene reconstruction. However, as the sequence length increases, these models suffer from catastrophic geometric forgetting and accumulation drift, primarily due to the quadratic complexity of global attention which necessitates truncated temporal windows. To overcome the resulting  geometric drift, we present Mamba-VGGT, an enhanced VGGT framework capable of persistent long-range reasoning. Our key contribution is a Sliding Window Mamba (SWM) memory module that maintains an explicit external memory token across temporal windows. This module leverages selective state-space modeling to distill and propagate global geometric priors, effectively bypassing the memory constraints of traditional transformers. To integrate these long-term temporal cues without disrupting the highly optimized spatial features of the pre-trained VGGT, we propose a Zero-Init Spatial Memory Injector. Utilizing zero-convolutional layers, this injector adaptively fuses persistent memory into the patch token stream, ensuring structural stability and seamless feature alignment. Extensive experiments  demonstrate that our approach significantly outperforms existing VGGT-based methods in maintaining spatial consistency and reducing trajectory accumulation errors. Our work provides a scalable, linear-complexity solution for geometry-grounded world modeling in extensive 3D environments.
\end{abstract}

\section{Introduction}
High-fidelity 3D scene reconstruction from monocular video is a cornerstone of computer vision, autonomous driving, robotics, and the development of general world representations~\cite{deng2025best3dscenerepresentation}. For decades, classical optimization frameworks dominated this landscape. They rely on computationally intensive offline processes and often falter on sparse or texture-less inputs. Recently, 3D foundation models, DUSt3R~\cite{wang2024dust3r}, VGGT~\cite{wang2025vggt}, $\pi^3$~\cite{wang2025pi} have emerged as a dominant paradigm by successfully lifting 2D visual observations into geometrically grounded 3D representations. By interleaving spatial and temporal attention, 3D foundation model frameworks effectively capture the intricate relationships between visual appearance and underlying geometry. However, despite their impressive performance on short video clips, these models face a fundamental bottleneck: the quadratic computational complexity of global attention. This constraint forces current architectures to operate on truncated temporal windows, leading to catastrophic geometric forgetting and accumulative trajectory drift when processing extensive video sequences.

The core challenge of scaling 3D foundation model to long-duration videos lies in the efficient propagation of geometric priors across distant frames. When the temporal context is limited, the model loses its "sense of history," failing to maintain global structural consistency, especially in scenarios involving large-scale loops or repetitive textures. While some works, such as CUT3R~\cite{cut3r}, Point3R~\cite{wu2025point3r}, TTT3R~\cite{ttt3r}, ZipMap~\cite{jin2026zipmap}, LoGER~\cite{zhang2026loger} attempt to address this through sparse attention or test-time training, they often sacrifice local reconstruction quality. Consequently, there is an urgent need for a memory mechanism that can maintain long-term persistence with linear complexity while remaining compatible with the highly optimized spatial grounding capabilities of pre-trained VGGTs.

In this paper, we propose Mamba-VGGT, a novel framework designed to empower Visual Geometry Grounded Transformers with persistent, long-range memory. Our key insight is to decouple the 3D spatial memory from the long-term temporal context. To achieve this, we introduce a Sliding Window Mamba (SWM) module that maintains and propagates an impilicit external memory token alongside the original patch token stream. By leveraging the selective state-space modeling (SSM) properties of Mamba, our SWM module distills geometric information from the current window and carries it into the next, ensuring a continuous and linear-time information flow throughout the entire video sequence. This architecture allows the model to "remember" distant geometric anchors without the prohibitive cost of global self-attention.

However, integrating such a dynamic external memory into a pre-trained VGGT, presents a non-trivial stability problem. Direct feature fusion can easily disrupt the delicate spatial feature distribution of the original transformer, leading to training divergence or degraded reconstruction quality. To address this, we design a Zero-Init Spatial Memory Injector based on the zero-convolution structure. This module serves as a "non-invasive" bridge, allowing the persistent memory from the Mamba stream to be adaptively and incrementally infused into the patch tokens. In the early stages of training, the zero-init layers ensure that the original VGGT's output remains unchanged, providing a stable foundation from which the model can gradually learn to leverage long-term temporal cues.

We evaluate our method on several datasets, from small indoor rooms to large-scale outdoor scenarios. Experimental results demonstrate that Mamba-VGGT significantly outperforms state-of-the-art VGGT-based approaches in maintaining structural integrity over long trajectories. Our approach not only reduces geometric drift  but also maintains a constant memory footprint relative to the sequence length. In summary, our contributions are three-fold:
\begin{itemize}
\item A novel framework centered on an external memory module is proposed to empower 3D foundation models in tackling long-sequence forgetting and accumulative trajectory drift.
    \item We introduce the Sliding Window Mamba (SWM) mechanism for VGGT, enabling linear-time long-range geometric memory propagation. We propose an explicit external memory token architecture that decouples temporal persistence from spatial feature extraction.
    \item We design a Zero-Init Spatial Memory Injector that ensures stable and effective integration of global priors into the pre-trained backbone. Experimental results on several datasets demonstrate the effectiveness of our method.
\end{itemize}

\section{Related Work}
\textbf{Learning-based visual SLAM}
Recent learning-based visual SLAM methods have demonstrated superior performance over classical approaches, such as ORB-SLAM3~\cite{campos2021orb}. These methods typically fall into two categories: those that learn robust 3D priors from large-scale datasets, such as DROID-SLAM~\cite{teed2021droid}, and those~\cite{zhu2022nice,Deng_2024_CVPR,deng2024compact,matsuki2024gaussian} that leverage implicit scene representations like NeRF and 3D Gaussian Splatting (3DGS) to achieve high-fidelity mapping. Despite these advancements, the quest for an optimal 3D scene representation~\cite{deng2025best3dscenerepresentation} for mapping and localization remains a fundamental challenge. With the emergence of 3D foundation models~\cite{wang2025vggt}, recent approaches~\cite{maggio2025vggt, deng2025vggt,shen2025grs} have begun to utilize powerful pretrained visual geometry models to resolve complex spatial dependencies. These frameworks rely on computationally expensive backends for graph construction, loop closure, and global bundle adjustment to ensure global consistency.

\textbf{Feed-forward 3D Reconstruction}
DUSt3R~\cite{wang2024dust3r} introduces a groundbreaking paradigm shift by directly regressing a pointmap from a pair of images without relying on any prior knowledge of the scene. Building upon this, MASt3R~\cite{leroy2024grounding} enhances the two-view prior to support more robust matching and tracking.
More recently, Some methods such as VGGT~\cite{wang2025vggt}, Fast3R~\cite{yang2025fast3r}, AMB3R~\cite{wang2025amb3r} FLARE~\cite{zhang2025flare}, $\pi^3$~\cite{wang2025pi}, and Lingbot-map~\cite{chen2026geometric}  have emerged as powerful foundation models for lifting 2D observations into consistent 3D representations. Despite its strengths in geometric representation, these frameworks are primarily optimized for short-to-medium sequences and struggles with long-sequence memory retention due to the quadratic complexity of its attention mechanism.
They often overlook spatial memory and multi-frame correlation, resulting in a lack of consistency during the mapping process. Our work addresses this limitation by introducing a Mamba-based sliding window memory into the VGGT architecture, providing a scalable and feedforward solution for maintaining global consistency across extensive temporal horizons.

\textbf{Memory for 3D foundation model}
Efficiently handling long sequences and memory propagation have motivated the development of linearcomplexity architectures, such as Linear Transformers~\cite{katharopoulos2020transformers}, Mamba~\cite{gu2023mamba} DeltaNet~\cite{schlag2021linear}, and Test-Time Training (TTT)~\cite{sun2024learning}. These models maintain compact recurrent states or employ online updates via gradient descent to capture extensive in-context information. Building on these foundations, several recent works have adapted such architectures for online 3D spatial memory. Spann3R~\cite{wang20253d} utilizes an external spatial memory for incremental reconstruction, while CUT3R~\cite{cut3r} incorporates recurrent states for sequential integration. \cite{fang2026incvggt, yuan2026infinitevggt} use top-k most relevant/highest-scoring slots for history KV cache. More recently, a surge of concurrent methods has utilized TTT layers to enhance long-sequence encoding; for instance, TTT3R refines recurrent states through test-time updates, and LaCT~\cite{Lact} dynamically updates non-linear MLP fast weights per token chunk. Further improvements in representation and scalability have been introduced by ZipMap\cite{jin2026zipmap}, LoGER~\cite{zhang2026loger}, Mem3R~\cite{liu2026mem3r}, Scal3r~\cite{xie2026scal3r}, and VGG-T$^3$~\cite{elflein2026vgg}. However, these TTT-based approaches rely on implicit memory encoding via fast-weight MLPs, which often leads to information loss in complex spatial representations. To address this, we propose a Mamba-based sliding window memory to better encode long-range spatial dependencies, coupled with a Zero-Init Spatial Injector to ensure the stable and lossless integration of memory into the network.

\begin{figure}[t]
  \centering
  \includegraphics[width=\linewidth]{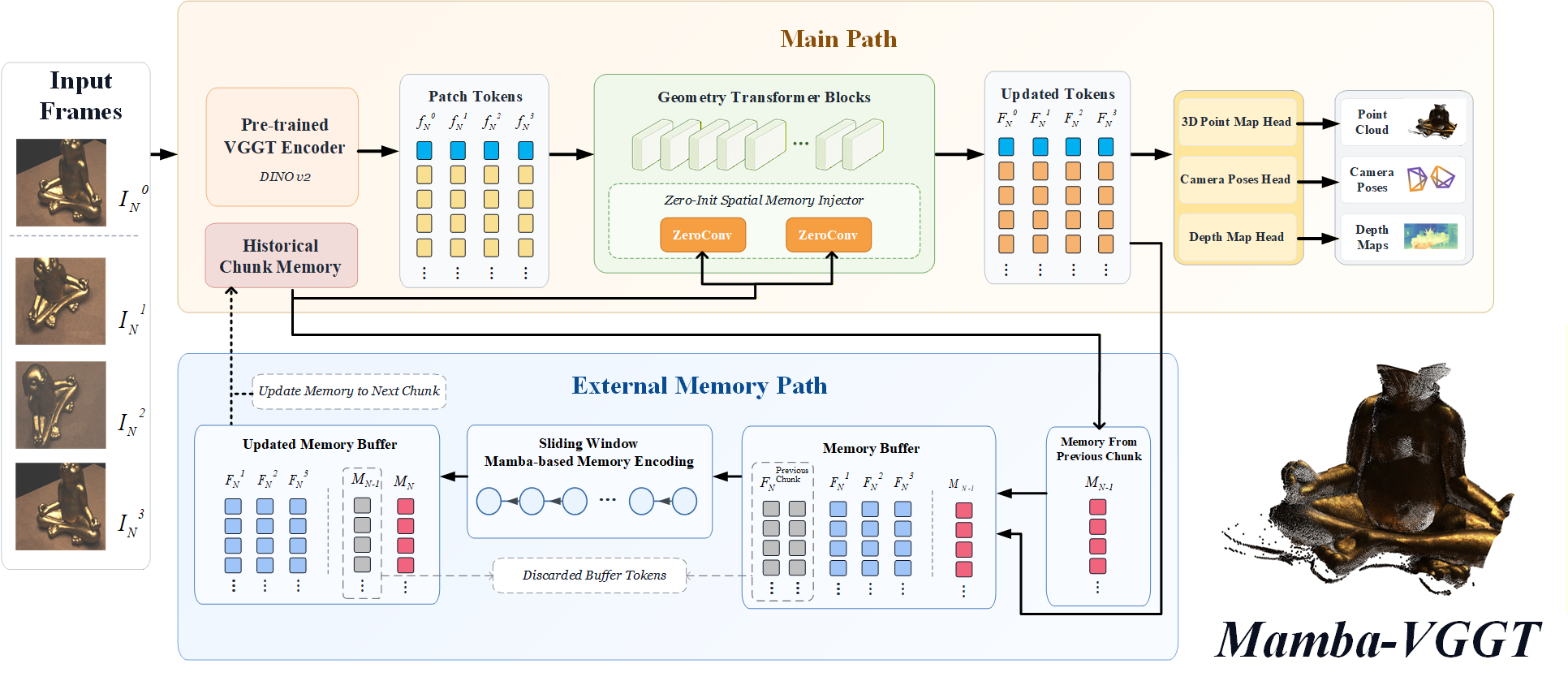}
  \caption{Overview of the Mamba-VGGT Architecture. The model takes long-duration video frames as input. Alongside the original patch token stream, we introduce a Sliding Window Mamba (SWM) module that maintains an explicit external memory token. To integrate long-term context without disrupting the pre-trained spatial features, a Zero-Init Spatial Memory Injector adaptively fuses the propagated memory back into the patch token stream. }
  \label{fig:framework}
\end{figure}

\section{Method}
The core philosophy of our framework is the 3D Foundation Model with a persistent, linear-time memory without compromising its established spatial grounding capabilities. As illustrated in Figure~\ref{fig:framework}, our architecture consists of two co-evolving streams: the Original Patch Token Stream for high-resolution geometry extraction and the External Memory Stream for long-term temporal reasoning.

\noindent \textbf{Input and Feature Representation} Given an input video sequence $\mathcal{V} = \{I_1, I_2, \dots, I_T\}$, we first partition it into a series of temporal windows $\mathcal{W} = \{W_1, \dots, W_K\}$. Each frame $I_t$ is tokenized into patch tokens $P_t \in \mathbb{R}^{N \times D}$ ($14 \times 14$ patches as specified in the VGGT backbone).

\noindent \textbf{The Tri-Module Execution Loop}
The framework operates through a continuous "Extract-Propagate-Inject" cycle across the sliding windows:

1. Backbone Feature Extraction: Given an input video sequence $\mathcal{V} = \{I_1, I_2, \dots, I_T\}$ consisting of $T$ frames, we first partition the sequence into a series of overlapping or adjacent temporal windows $\mathcal{W} = \{W_1, W_2, \dots, W_K\}$, where each window contains $L$ frames. Following the VGGT paradigm, each frame $I_t$ is tokenized into a set of patch tokens $P_t \in \mathbb{R}^{N \times D}$, where $N$ is the number of patches (e.g., $14 \times 14$ tokens) and $D$ is the embedding dimension. This backbone is highly efficient, supporting linear-time, bidirectional reconstruction of camera poses $\{c_1, \dots, c_T\}$, depth maps $\{D_1, \dots, D_T\}$, and point clouds $\{p_1, \dots, p_T\}$, all in a single feed-forward pass. While this allows for rapid local grounding, our framework extends its capability to maintain this geometric accuracy over much larger temporal scales by addressing the window-to-window drift.

2. External Memory Propagation (SWM): Parallel to the backbone, our Sliding Window Mamba (SWM) module maintains an explicit External Memory Token $M_k$. This token distills the geometric essence of the current window $W_k$ and updates its hidden state $h_k$ via selective state-space modeling. The updated memory is then propagated to the subsequent window $W_{k+1}$, carrying a "geometric summary" across long temporal horizons that a standard transformer window would otherwise forget.

3. Spatial Information Injection (Zero-Init Injector): The distilled memory $M_k$ is actively fed back into the VGGT's transformer layers through our Zero-Init Spatial Memory Injector. Utilizing zero-convolutional layers, the injector adaptively aligns and fuses the global temporal priors from $M_k$ back into the local patch tokens $P_t$. This ensures that the generated point clouds and poses remain globally consistent even during extended trajectories.

\subsection{Persistent Memory Propagation via Selective Mamba Modeling}
To overcome the temporal limitations of the standard Transformer architecture, we introduce a dedicated memory stream that runs parallel to the VGGT backbone. The goal is to compress the geometric information of each sliding window into a compact External Memory Token and propagate it across the entire video sequence using a Mamba-based State-Space Model~\cite{gu2023mamba}.

\noindent \textbf{Dual-Stream Memory Buffers} Unlike previous registration works that rely on pairwise frames, our framework processes continuous point cloud video clips. We define two fixed-length buffers with a temporal horizon of $T$: K Buffer $M^K \in \mathbb{R}^{T \times D_f}$ and V Buffer $M^V \in \mathbb{R}^{T \times D_p}$: Stores the latent geometric features extracted from the patch tokens. Both buffers are initialized as empty sets and follow a sliding window update mechanism to maintain a persistent "geometric history" as new frames are delivered.

\textbf{Window-based Memory Read-out} When a new window of frames is processed at time $t$, we perform a Memory Read-out operation to synthesize the current features $F_t$ (derived from the VGGT patch tokens) with the temporally-stored history. Following the sliding window logic, the temporally-farthest feature in the buffer $\hat{F}_{t-T-1}$ is discarded. The remaining $T-1$ nearest features are concatenated with the current feature $F_t$ to formulate the input tokens for temporal encoding:
\begin{equation}
    M_{t-1}^F : \{\hat{F}_{t-T}, \dots, \hat{F}_{t-1}, F_t\} 
\end{equation}
where $\{\}$ indicates concatenation along the temporal dimension. This ensures that the subsequent Mamba block has access to a continuous stream of spatio-temporal context.

\textbf{Mamba-based Temporal Encoding} The concatenated features in $M_{t-1}^F$ are fed into a Mamba-based temporal encoding block to capture long-term dependencies with linear complexity. The encoding process is governed by the following equations:
\begin{align}
     \hat{M}_{t-1}^F &= \text{LN}(M_{t-1}^F) \\
     \overline{M}_t^F &= \sigma(\text{DW}(\text{Linear}(\hat{M}_{t-1}^F))) \\
     \hat{M}_t^F &= \sigma(\text{Linear}(\hat{M}_{t-1}^F)) \\
     \hat{F}_t &= \text{Linear}(\text{SSM}(\overline{M}_t^F)) \odot \hat{M}_t^F + M_{t-1}^F
\end{align}
where $\text{LN}$ denotes Layer Normalization, $\text{DW}$ is Depth-Wise convolution, and $\sigma$ represents the SiLU activation function. The Selective State-Space Model (SSM) serves as the core reasoning engine, distilling the long-term temporal features into the refined output $\hat{F}_t$.

\noindent \textbf{Memory Update Mechanism} 
To ensure the model adapts to the evolving scene, the buffers are progressively updated in a first-in-first-out (FIFO) manner. Upon completing the encoding for the current frame, the newly encoded feature $\hat{F}_t$ is appended to the buffer, while the oldest entry is removed:$$M_t^F = \{\hat{F}_{t-T+1}, \dots, \hat{F}_{t-1}, \hat{F}_t\}$$This persistent update cycle allows the 3D foundation model to maintain a constant-size memory footprint while theoretically accessing an infinite temporal horizon through the recursive nature of the Mamba state.

\begin{figure}[t]
  \centering
  \includegraphics[width=\linewidth]{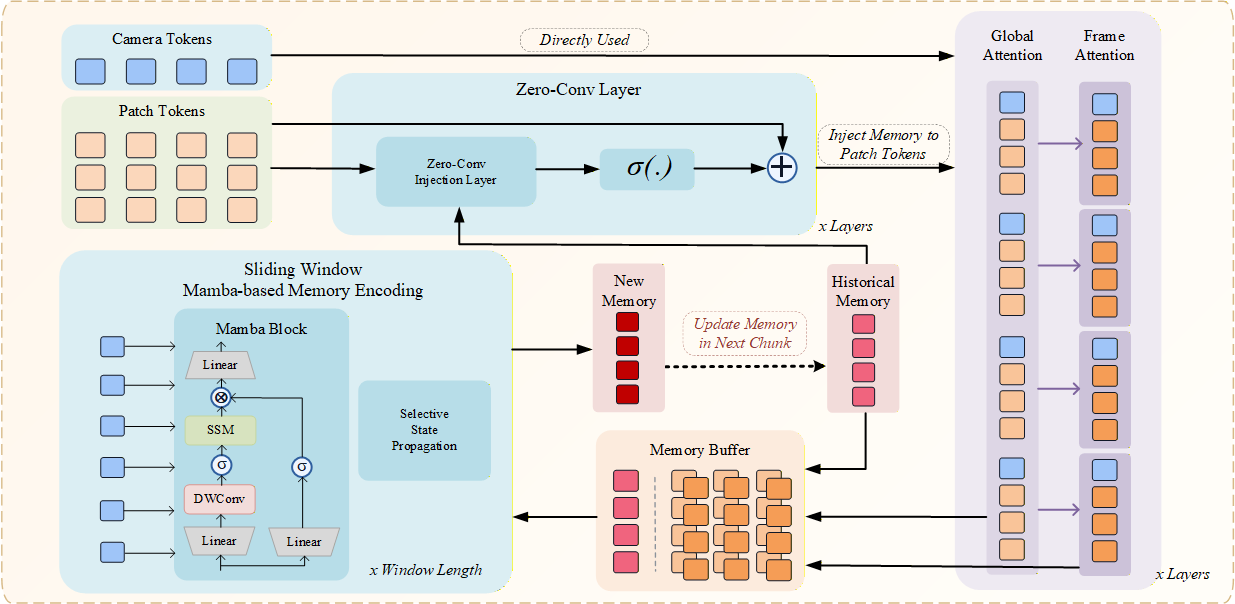}
  \caption{Detailed Architecture of the Mamba-VGGT with Sliding Window Memory and Zero-Conv block. We present the memory read-out process executed by the Zero-Init Spatial Memory Injector, which adaptively retrieves distilled temporal priors from the Mamba state and fuses them back into the spatial stream.}
  \label{fig:framework2}
\end{figure}

\subsection{Spatial Information Injection via Zero-Init Injector}
While the Mamba-based temporal stream effectively captures long-term geometric dependencies, integrating these temporal priors back into a pre-trained 3D Foundation Model presents a significant challenge. Direct feature fusion or summation often disrupts the highly optimized spatial feature distribution of VGGT, leading to training instability or forgetting of the backbone's grounding capabilities. To bridge this gap, we design a Zero-Init Spatial Memory Injector that serves as a non-invasive, adaptive bridge between temporal memory and spatial reconstruction.

The core of our injector is the Zero-Convolution (Zero-Conv) structure, which is a $1 \times 1$ convolutional layer  whose weights and biases are initialized to zero. For a given patch $i$ (within the $14 \times 14$ token grid) at time $t$, let $K_{i,t}$ and $V_{i,t}$ be the original key and value tokens from the VGGT backbone, and let $\hat{K}_{i,t}$ and $\hat{V}_{i,t}$ be the temporally-refined tokens output by the Mamba memory stream. The injected KV tokens are formulated as:
\begin{align}
    K'_{i,t} = K_{i,t} + \text{ZeroConv}(\hat{K}_{i,t}; \Theta_K) \\
    V'_{i,t} = V_{i,t} + \text{ZeroConv}(\hat{V}_{i,t}; \Theta_V) 
\end{align}
where $\Theta_K$ and $\Theta_V$ are the zero-initialized parameters. This design ensures that at the beginning of training, the injection module outputs exactly zero, allowing the Visual Geometry Grounded Transformer to operate in its original state. This "cold-start" protection is crucial for preserving the pre-trained geometric knowledge while the Mamba module begins to learn the complex temporal correlations.

As training progresses, the Zero-Init Injector adaptively learns to weight the importance of the long-term memory. In regions with significant camera motion or sparse visual features, the model learns to increase the influence of $\hat{K}$ and $\hat{V}$ to maintain structural integrity. The injected tokens $\{K', V'\}$ are then used in the standard attention mechanism of the subsequent transformer layers:
\begin{equation}
    \text{Attention}(Q_t, K'_t, V'_t) = \text{Softmax}\left(\frac{Q_t (K'_t)^T}{\sqrt{d_k}}\right) V'_t 
\end{equation}
By grounding the current query $Q_t$ against the memory-augmented keys and values, the framework achieves a persistent spatial grounding. This mechanism allows the model to "re-observe" historical geometric anchors and effectively eliminate the accumulation drift that typically occurs when a 3D foundation model is restricted to short-term temporal windows.

\subsection{Training and Optimization Strategy}
To preserve the powerful spatial grounding capabilities of the pre-trained 3D Foundation Model, we adopt a decoupled multi-stage training strategy. This approach ensures that the framework progressively learns to balance high-resolution geometric precision with long-term temporal consistency.

The first stage is parameter-efficient backend warm-up. We freeze the front-end model and only train the Sliding Window Mamba (SWM) blocks and the Zero-Init Spatial Memory Injectors. This strategy not only reduces the computational burden but also prevents the catastrophic forgetting of the backbone's pre-trained geometric knowledge. Our optimization objective follows the multi-task loss structure of VGGT for pointmap, depth, and camera pose estimation:
\begin{equation}
    \mathcal{L} = \mathcal{L}_{\text{depth}} + \mathcal{L}_{\text{pointmap}} + \mathcal{L}_{\text{camera}}
\end{equation}
This warm-up phase allows the Mamba-based memory stream to effectively learn the compression and propagation of KV caches without disrupting the stable feature distribution of the backbone.

The second stage is global joint fine-tuning and scaling. During this stage, we strategically increase the input sequence length and expand the temporal window size. This "long-context" training regime is designed to enhance the framework's scalability and its ability to handle extensive trajectories. By exposing the model to more complex temporal dependencies and larger-scale scene structures, we empower the network to leverage the full capacity of the Mamba hidden states, effectively eliminating accumulative drift and ensuring global geometric coherence across "infinite" video sequences.

\begin{table*}[t]
\centering
\caption{Reconstruction evaluation on DTU~\cite{dtu} and ETH3D~\cite{eth3d} under the streaming-input setting.}
\resizebox{\textwidth}{!}{
\begin{tabular}{lcccccccccccc}
\toprule
\multirow{3}{*}{Method}
& \multicolumn{6}{c}{DTU~\cite{dtu}}
& \multicolumn{6}{c}{ETH3D~\cite{eth3d}} \\
\cmidrule(lr){2-7}
\cmidrule(lr){8-13}
& \multicolumn{2}{c}{Acc. $\downarrow$}
& \multicolumn{2}{c}{Comp. $\downarrow$}
& \multicolumn{2}{c}{NC. $\uparrow$}
& \multicolumn{2}{c}{Acc. $\downarrow$}
& \multicolumn{2}{c}{Comp. $\downarrow$}
& \multicolumn{2}{c}{NC. $\uparrow$} \\
\cmidrule(lr){2-3}
\cmidrule(lr){4-5}
\cmidrule(lr){6-7}
\cmidrule(lr){8-9}
\cmidrule(lr){10-11}
\cmidrule(lr){12-13}
& Mean & Med.
& Mean & Med. 
& Mean & Med.
& Mean & Med.
& Mean & Med. 
& Mean & Med. \\
\midrule
Fast3R~\cite{yang2025fast3r}
& 4.540 & 3.919 & 2.929 & 2.125 & 0.671 & 0.755
& 0.832 & 0.691 & 0.978 & 0.683 & 0.667 & 0.766 \\
FLARE~\cite{zhang2025flare}
& 4.541 & 3.468 & 3.174 & 2.420 & 0.684 & 0.774
& 0.764 & 0.638 & 0.964 & 0.695 & 0.744 & 0.864 \\
TTT3R~\cite{ttt3r}
& 5.337 & 3.261 & 6.593 & 4.236 & 0.666 & 0.743
& 0.763 & 0.633 & 0.881 & 0.617 & 0.739 & 0.840 \\
CUT3R~\cite{cut3r}
& 5.045 & 2.954 & 6.437 & 4.146 & 0.666 & 0.742
& 0.593 & 0.461 & 0.747 & 0.590 & 0.754 & 0.863 \\
VGGT(windowed)~\cite{wang2025vggt}
& 5.031 & 3.017 & 5.017 & 3.533 & 0.782 & 0.837
& 0.561 & 0.439 & 0.665 & 0.381 & \textbf{0.883} & \textbf{0.942} \\

ZipMap(stream)~\cite{jin2026zipmap}
& 4.107 & 2.682 & \textbf{3.492} & \textbf{2.147} & 0.695 & 0.771
& 0.609 & 0.487 & 0.941 & 0.657 & 0.756 & 0.857 \\
\textbf{Ours}
& \textbf{4.065} & \textbf{2.553} & 3.962 & 2.271 & \textbf{0.784} & \textbf{0.839}
& \textbf{0.541} & \textbf{0.323} & \textbf{0.639} & \textbf{0.364} & 0.861 & 0.936 \\
\bottomrule
\end{tabular}
}
\label{tab:reconstruction}
\vspace{-0.6cm}
\end{table*}

\begin{table*}[t]
\centering
\caption{Video depth evaluation under the streaming-input setting. }
\label{tab:video_depth_eval}
\begin{tabular}{lcccc}
\toprule
\multirow{2}{*}{Method}
& \multicolumn{2}{c}{Sintel~\cite{butler2012naturalistic}}
& \multicolumn{2}{c}{Bonn~\cite{bonn}} \\
\cmidrule(lr){2-3}
\cmidrule(lr){4-5}
& AbsRel $\downarrow$
& $\delta < 1.25$ $\uparrow$
& AbsRel $\downarrow$
& $\delta < 1.25$ $\uparrow$ \\
\midrule
Spann3R~\cite{wang20253d}
& 0.622 & 0.426
& 0.144 & 0.813 \\
Fast3R~\cite{yang2025fast3r}
& 0.638 & 0.422
& 0.194 & 0.772 \\
FLARE~\cite{zhang2025flare}
& 0.729 & 0.336
& 0.152 & 0.790 \\
CUT3R~\cite{cut3r}
& 0.432 & 0.510
& 0.072 & 0.951 \\
TTT3R~\cite{ttt3r}
& 0.426 & 0.522
& 0.061 & 0.970 \\
StreamVGGT~\cite{zhuo2025streaming}
& 0.363 & 0.607
& 0.069 & 0.961 \\
ZipMap(stream)~\cite{jin2026zipmap}
& \textbf{0.334} & 0.612
& 0.056 & 0.973 \\
VGGT (Windowed)~\cite{wang2025vggt}
& 0.524 & 0.456
& 0.069 & 0.954 \\
\textbf{Ours}
& 0.354 & \textbf{0.618}
& \textbf{0.055} & \textbf{0.977} \\
\bottomrule
\end{tabular}
\label{tab:depth}
\vspace{-0.6cm}
\end{table*}

\begin{table*}[t]
\centering
\caption{Camera pose evaluation under the streaming-input setting.}
\label{tab:camera_pose_eval}
\resizebox{\textwidth}{!}{
\begin{tabular}{lcccccc}
\toprule
\multirow{2}{*}{Method}
& \multicolumn{3}{c}{RealEstate10K~\cite{RealEstate10K}}
& \multicolumn{3}{c}{Co3Dv2~\cite{Co3Dv2}} \\
\cmidrule(lr){2-4}
\cmidrule(lr){5-7}
& AUC @ 5
& AUC @ 15
& AUC @ 30
& AUC @ 5
& AUC @ 15
& AUC @ 30\\ \midrule
Fast3R~\cite{yang2025fast3r}
& 22.36 & 46.71 & 61.68
& 31.05 & 59.63 & 73.43 \\
FLARE~\cite{zhang2025flare}
& 38.47 & 67.02 & 80.01
& 23.84 & 57.78 & 73.99 \\
CUT3R~\cite{cut3r}
& 46.92 & 70.65 & 81.68
& 24.88 & 56.28 & 71.72 \\
TTT3R~\cite{ttt3r}
& 46.37 & 70.33 & 81.51
& 22.61 & 53.49 & 69.46\\
ZipMap(Stream)~\cite{jin2026zipmap}
& 28.17 & 58.36 & 73.24
& 45.38 & 72.58 & 83.12\\
VGGT(Windowed)~\cite{wang2025vggt}
& 37.42 & 64.91 & 81.67
& \textbf{64.43} & 81.31 & \textbf{87.92}\\
\textbf{Ours}
& \textbf{47.13} & \textbf{71.58} & \textbf{83.26}
& 58.87 & \textbf{82.02} & 85.44 \\
\bottomrule
\end{tabular}
}
\vspace{-0.6cm}
\end{table*}

\begin{table*}[t]
\centering
\caption{Reconstruction evaluation on 7-Scenes}
\label{tab:recon_7scenes_nrgbd}
\resizebox{\textwidth}{!}{
\begin{tabular}{lcccccccccccc}
\toprule
\multirow{3}{*}{Method}
& \multicolumn{6}{c}{7-Scenes(Sparse)}
& \multicolumn{6}{c}{7-Scenes(Dense)} \\
\cmidrule(lr){2-7}
\cmidrule(lr){8-13}
& \multicolumn{2}{c}{Acc.$\downarrow$}
& \multicolumn{2}{c}{Comp.$\downarrow$}
& \multicolumn{2}{c}{NC.$\uparrow$}
& \multicolumn{2}{c}{Acc.$\downarrow$}
& \multicolumn{2}{c}{Comp.$\downarrow$}
& \multicolumn{2}{c}{NC.$\uparrow$} \\
\cmidrule(lr){2-3}
\cmidrule(lr){4-5}
\cmidrule(lr){6-7}
\cmidrule(lr){8-9}
\cmidrule(lr){10-11}
\cmidrule(lr){12-13}
& Mean & Med.
& Mean & Med.
& Mean & Med.
& Mean & Med.
& Mean & Med.
& Mean & Med. \\
\midrule
Fast3R~\cite{yang2025fast3r}
& 0.095 & 0.065 & 0.144 & 0.089 & 0.673 & 0.759
& 0.040 & 0.017 & 0.056 & 0.018 & 0.644 & 0.725 \\
CUT3R~\cite{cut3r}
& 0.080 & 0.055 & 0.102 & 0.066 & 0.711 & 0.811
& 0.023 & 0.010 & 0.028 & 0.008 & 0.674 & 0.771 \\
TTT3R~\cite{ttt3r}
& 0.098 & 0.062 & 0.159 & 0.107 & 0.681 & 0.768
&  0.035 & 0.016 & 0.032 & 0.010 & 0.666 & 0.760 \\
FLARE~\cite{zhang2025flare}
& 0.085 & 0.057 & 0.145 & 0.107 & 0.696 & 0.780
& \textbf{0.019} & \textbf{0.007} & 0.026 & 0.013 & 0.684 & 0.785 \\
VGGT(windowed)~\cite{wang2025vggt}
& 0.057 & 0.036 & 0.064 & 0.033 & 0.733 & 0.846
& 0.021 & 0.015 & 0.032 & 0.019 & 0.667 & 0.760 \\
Pi$^3$(windowed)~\cite{wang2025pi}  & 0.053 & 0.030 & 0.062 & 0.033 & 0.771 & 0.854
& \textbf{0.019} & \textbf{0.007} & \textbf{0.024} & \textbf{0.012} & 0.683 & 0.7683 \\
ZipMap(streaming)~\cite{jin2026zipmap}
& 0.053 & \textbf{0.028} & 0.076 & 0.051 & 0.741 & 0.840
& 0.021 & 0.013 & 0.032 & 0.018 & 0.680 & 0.780 \\
VGGT-SLAM~\cite{maggio2025vggt} & 0.054 & 0.032 & 0.063 & 0.036 & 0.774 & 0.881
& 0.021 & 0.013 & 0.032 & 0.016 & 0.671 & 0.772 \\
\textbf{Ours}
& \textbf{0.051} & 0.030 & \textbf{0.061} & \textbf{0.034} & \textbf{0.775} & \textbf{0.886}
& \textbf{0.019} & 0.009 & 0.028 & 0.014 & \textbf{0.685} & \textbf{0.788} \\
\bottomrule
\end{tabular}
}
\vspace{-0.6cm}
\end{table*}

\begin{figure}[t]
  \centering
  \includegraphics[width=\linewidth]{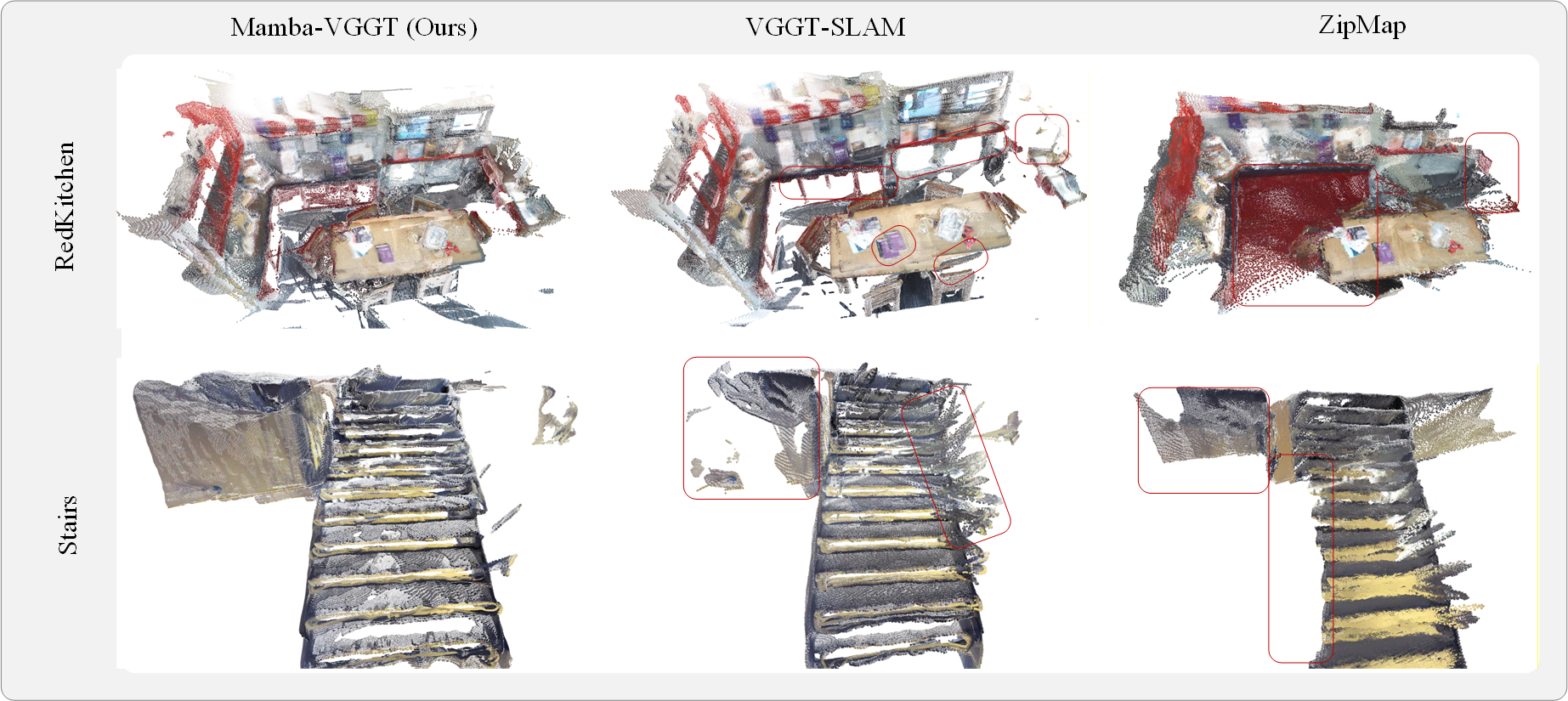}
  \caption{We present the qualitative results of our method and other baseline on long sequence reconstruction.}
  \label{fig:reconstrction}
  \vspace{-0.3cm}
\end{figure}

\begin{figure}[t]
  \centering
  \includegraphics[width=\linewidth]{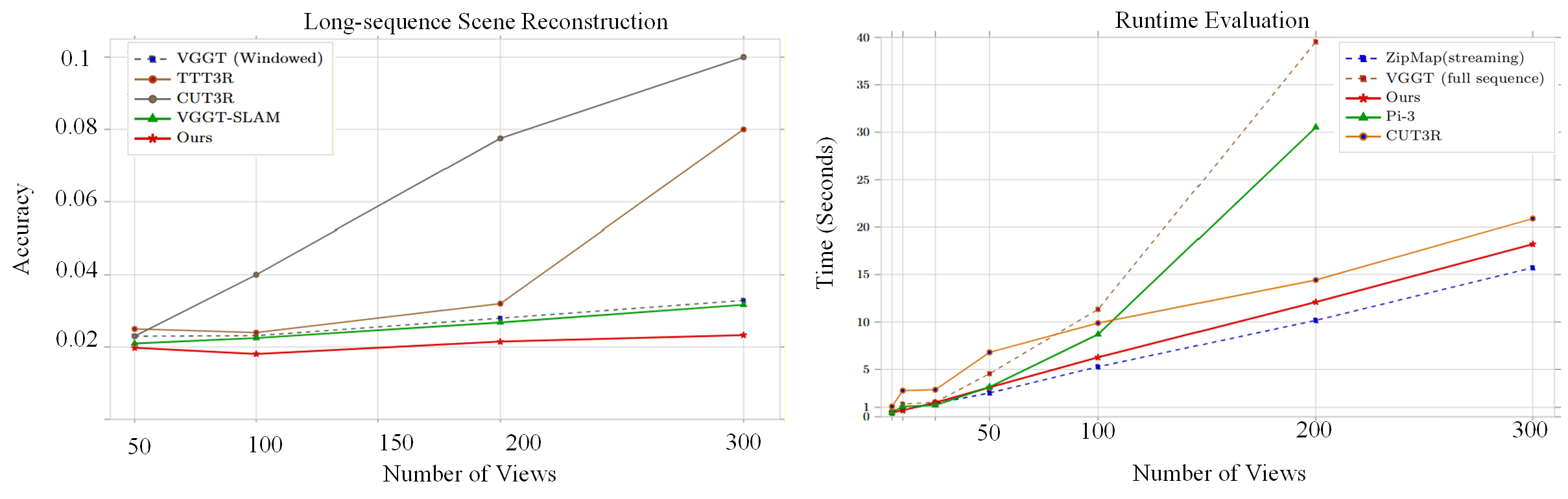}
  \caption{Performance analysis across sequence lengths. (a) long-sequence scene reconstruction results and (b)  Runtime evaluation  with different input frame count. }
  \label{fig:runtime}
  \vspace{-0.3cm}
\end{figure}

\begin{table*}[t]
\centering
\caption{Ablation of reconstruction performance on 7-Scenes with different mamba update parameters.}
\begin{tabular}{lcccccc}
\toprule
\multirow{3}{*}{Method}
& \multicolumn{6}{c}{7-Scenes } \\
\cmidrule(lr){2-7}
& \multicolumn{2}{c}{Acc. $\downarrow$}
& \multicolumn{2}{c}{Comp. $\downarrow$}
& \multicolumn{2}{c}{NC $\uparrow$} \\
\cmidrule(lr){2-3}
\cmidrule(lr){4-5}
\cmidrule(lr){6-7}
& Mean & Med.
& Mean & Med.
& Mean & Med. \\
\midrule
w/o mamba update($\alpha = 0$)
& 0.021 & 0.011 & 0.032 & 0.016 & 0.679 & 0.773 \\
w/o memory
& 0.024 & 0.012 & 0.034 & 0.018 & 0.661 & 0.757 \\
w/o Zero-Init Injector
& 0.028 & 0.013 & 0.033 & 0.017 & 0.667 & 0.766 \\
Ours
& \textbf{0.019} & \textbf{0.009} & \textbf{0.028} & \textbf{0.014} & \textbf{0.685} & \textbf{0.788} \\
\bottomrule
\end{tabular}
\label{tab:ablation}
\vspace{-0.6cm}
\end{table*}

\subsection{Potential for Large-Scale SLAM Framework}
The proposed framework provides a robust foundation for large-scale SLAM systems. By synergizing the sub-map abstraction of VGGT-SLAM~\cite{maggio2025vggt} with our Mamba-driven persistent memory, we pave the way for a SLAM architecture that excels in both local precision and global consistency.

\noindent\textbf{From Local Sub-maps to Sequential Continuity} 
Conventional learning-based SLAM systems often struggle with the isolation of sub-maps, where truncated temporal windows lead to accumulated drift over long trajectories. Our framework addresses this by treating each sub-map as a temporal unit within a continuous stream. Mamba block serves as the core distillation engine, extracting a "geometric summary" from the patch tokens of the current sub-map and propagating this memory to the subsequent one. This ensures that the system maintains a persistent global state, allowing the current camera pose and scene geometry to be grounded against the entire historical trajectory rather than just a few preceding frames. 

\noindent\textbf{Enhanced Backend Optimization via SL4}
A unique advantage of our framework is its seamless integration with the SL4 optimization backend of VGGT-SLAM. In our proposed SLAM pipeline, the temporally refined memory tokens infused back into the patch tokens via the Zero-Init Injector, providing a significantly more accurate prior for backend refinement. By supporting serialized video input with linear-time complexity, our framework transforms VGGT-based models from local reconstructors into scalable world-modeling engines.

\vspace{-0.4cm}
\section{Experiments}
We evaluate Mamba-VGGT on a comprehensive suite of 3D tasks, including camera pose estimation, point-map reconstruction, and video depth estimation. Our evaluation is organized into three complementary
settings: comparison with streaming-input methods, comparison with non-streaming methods, and SLAM framework evaluation.

For streaming-input evaluation, we follow the metric design used by ZipMap. We compare our method
with the official online streaming reconstruction variant of ZipMap and reuse part of the results reported
in its benchmark. We also construct a Windowed VGGT baseline by applying the same window splitting
and first-frame anchoring strategy used in our model to VGGT. For fairness, no method uses additional
post-alignment.

For SLAM framework evaluation, we integrate our model into the VGGT-SLAM framework. We follow
its overlap-frame estimation and SL4 submap alignment pipeline to evaluate model performance under
long sequential input. The results show that the proposed Mamba-based memory generation and injection
mechanism effectively improves the long-sequence reconstruction ability of VGGT under a finite-window setting.

\noindent\textbf{Implementation details.}
We train our model on a mixture of real and synthetic datasets. Detailed dataset
information is provided in Appendix. The training process are conducted on NVIDIA A100 GPUs. The backbone uses the pretrained VGGT image encoder based on DINOv2, and the original VGGT aggregation blocks are initialized from pretrained VGGT parameters. The token dimension is set to 1024. Each zero-conv injection branch consists of two 1x1 Conv1d layers with a GELU activation in between. This makes the memory branch initially equivalent to an identity-preserving perturbation and allows the model to gradually learn how much memory should affect the pretrained VGGT representation.

\subsection{Benchmark Evaluation}

\noindent\textbf{Point-Map Estimation}
We evaluate point-map reconstruction on DTU~\cite{dtu} and ETH3D~\cite{eth3d}. Following the
ZipMap-style reconstruction protocol, we compare our method with the streaming version of ZipMap, its
reported benchmark baselines, and our Windowed VGGT baseline. We use accuracy, completeness, and
normal consistency as evaluation metrics. Tab~\ref{tab:reconstruction} summarizes the streaming-input comparison. Under the same streaming setting, our method outperforms ZipMap in accuracy and normal consistency,
 and completeness varies across datasets. This indicates that the proposed memory mechanism improves
local geometric precision and surface orientation consistency.

\textbf{Video Depth Estimation}
We follow the ZipMap setting for video depth evaluation on Sintel~\cite{butler2012naturalistic} and Bonn~\cite{bonn}.
We report standard depth metrics under the same alignment protocol used in the ZipMap benchmark.
Table~\ref{tab:depth} compares our method with streaming-input methods. 

\textbf{Camera Pose Estimation}
We first evaluate camera pose estimation on RealEstate10K~\cite{RealEstate10K} and Co3Dv2~\cite{Co3Dv2}.
We report pose AUC under angular error thresholds of 5, 15, and 30 degrees. Table~\ref{tab:camera_pose_eval} compares our method
with streaming baselines. 
Under the same streaming-input setting, our method consistently outperforms ZipMap, showing that the proposed memory mechanism provides a
clear advantage over a simple windowed adaptation of VGGT. 

\vspace{-0.3cm}
\subsection{Large-scale SLAM Framework Evaluation}
For long-sequence reconstruction, we compare VGGT-SLAM and our
model integrated into the same VGGT-SLAM framework. Our model demonstrates a clear advantage as the sequence length increases. By leveraging the Mamba-based external memory, our framework effectively mitigates the catastrophic forgetting of spatial anchors, resulting in significantly higher structural accuracy and lower geometric drift compared to the baseline. The trend in Fig.~\ref{fig:runtime} highlights a key strength of our approach: while the baseline's performance degrades as the temporal horizon expands, our model's reconstruction quality remains robust. This confirms that our Sliding Window Mamba effectively propagates geometric priors, enabling the 3D Foundation Model to scale to extensive trajectories with linear-time complexity and constant memory overhead.

\vspace{-0.3cm}
\subsection{Efficiency and Scalability}
We further evaluate the efficiency and scalability of our model under streaming and long-sequence settings. Fig.~\ref{fig:runtime} compares the long-sequence runtime of ZipMap, VGGT-SLAM, and our proposed method as a function of the number of input views. Even as the number of views scales from 50 to 300 and beyond, the runtime of our model  is better than other baselines. This result highlights that our decoupled architecture, which separates high-resolution spatial grounding from linear-time temporal propagation, provides a highly scalable solution for large-scale 3D world modeling without the time bottlenecks associated with global attention mechanisms.

\vspace{-0.3cm}
\subsection{Ablation Study}
\vspace{-0.3cm}
We conduct comprehensive ablation studies to validate the effectiveness of our persistent memory module and the zero-init injection strategy. To verify the necessity of the memory stream, we compare our full model against a baseline where the memory module is removed. As shown in Tab.\ref{tab:ablation}, the absence of memory and memory update the leads to a significant drop in reconstruction completeness for sequences longer than 100 frames. This confirms that our Mamba-based memory is the primary driver for overcoming the "short-term bias" of the foundation model. We further evaluate the importance of the Zero-Conv structure by replacing it with a standard linear projection (randomly initialized). We observe that without zero-initialization, the model suffers from training instability in the first stage, often leading to distorted point clouds as the new memory signals "shock" the pre-trained spatial features.
\vspace{-0.3cm}
\section{Conclusion}
\vspace{-0.3cm}
We present Mamba-VGGT, a framework that scales Visual Geometry Grounded Transformers to long-duration video sequences by addressing the quadratic complexity of global attention. Our approach reformulates the static KV cache into a persistent memory stream via Sliding Window Mamba (SWM), enabling linear-time propagation of geometric priors and mitigating accumulative trajectory drift. Through the Zero-Init Spatial Memory Injector, we achieve stable, non-invasive integration of temporal context without disrupting the backbone’s pre-trained spatial grounding. Experimental results confirm that Mamba-VGGT maintains state-of-the-art accuracy with a constant memory footprint, providing a scalable and efficient solution for consistent 3D world modeling in extensive environments and the potential ability for large-scale SLAM framework.

\bibliographystyle{plain}
\bibliography{main.bib}

\end{document}